# Dynamic Retrieval Augmented Generation of Ontologies using Artificial Intelligence (DRAGON-AI)

Sabrina Toro[1], Anna V Anagnostopoulos[2], Susan M Bello[2], Kai Blumberg[3], Rhiannon Cameron[4], Leigh Carmody[5], Alexander D Diehl[6], Damion M Dooley[4], William D Duncan[7], Petra Fey[8], Pascale Gaudet[9], Nomi L Harris[10], Marcin P Joachimiak[10], Leila Kiani[11], Tiago Lubiana[12], Monica C Munoz-Torres[13], Shawn O'Neil[1], David Osumi-Sutherland[14], Aleix Puig-Barbe[15], Justin T Reese[10], Leonore Reiser[16], Sofia MC Robb[17], Troy Ruemping[18], James Seager[19], Eric Sid[20], Ray Stefancsik[15], Magalie Weber[21], Valerie Wood[22], Melissa A Haendel[1], Christopher J Mungall[10,*]

[1]University of North Carolina at Chapel Hill, Chapel Hill, NC, USA, [2]The Jackson Laboratory, Bar Harbor, ME, USA, [3]University of Arizona, Tucson, AZ, USA, [4]Simon Fraser University, Burnaby, BC, Canada, [5]The Jackson Laboratory for Genomic Medicine, Farmington, CT, USA, [6]University at Buffalo, Buffalo, NY, USA, [7]University of Florida, Gainsville, FL, USA, [8]Northwestern University, Evanston, IL, USA, [9]SIB Swiss Institute of Bioinformatics, Geneva, Switzerland, [10]Lawrence Berkeley National Laboratory, Berkeley, CA, USA, [11]Independent Scientific Information Analyst, [12]University of São Paulo, São Paulo, Brazil, [13]University of Colorado Anschutz Medical Campus, Aurora, CO, USA, [14]Sanger Institute, Hinxton, UK, [15]European Bioinformatics Institute (EMBL-EBI), Hinxton, UK, [16]Phoenix Bioinformatics, Newark, CA, USA, [17]Stowers Institute for Medical Research, Kansas City, MO, USA, [18]IC-FOODS, Austin, TX, USA, [19]Rothamsted Research, Harpenden, UK, [20]National Center for Advancing Translational Sciences, Bethesda, MD, USA, [21]INRAE, French National Research Institute for Agriculture, Food and Environment, UR BIA, Nantes, France, [22]University of Cambridge, Cambridge, UK

* corresponding author <cjmungall@lbl.gov>

# Abstract

## Background

Ontologies are fundamental components of informatics infrastructure in domains such as biomedical, environmental, and food sciences, representing consensus knowledge in an accurate and computable form. However, their construction and maintenance demand substantial resources and necessitate substantial collaboration between domain experts, curators, and ontology experts.

We present Dynamic Retrieval Augmented Generation of Ontologies using AI (DRAGON-AI), an ontology generation method employing Large Language Models (LLMs) and Retrieval Augmented Generation (RAG). DRAGON-AI can generate textual and logical ontology components, drawing from existing knowledge in multiple ontologies and unstructured text sources.

## Results

We assessed performance of DRAGON-AI on de novo term construction across ten diverse ontologies, making use of extensive manual evaluation of results. Our method has high precision for relationship generation, but has slightly lower precision than from logic-based reasoning. Our method is also able to generate definitions deemed acceptable by expert evaluators, but these scored worse than human-authored definitions. Notably, evaluators with the highest level of confidence in a domain were better able to discern flaws in AI-generated definitions. We also demonstrated the ability of DRAGON-AI to incorporate natural language instructions in the form of GitHub issues.

## Conclusions

These findings suggest DRAGON-AI's potential to substantially aid the manual ontology construction process. However, our results also underscore the importance of having expert curators and ontology editors drive the ontology generation process.

# Background

Ontologies are structured representations of knowledge, consisting of a collection of terms organized using logical relationships and textual information. In the life sciences, ontologies such as the Gene Ontology (GO) (1), Mondo (2), Uberon (3), and FoodON (4) are used for a variety of purposes such as curation of gene function and expression, classification of diseases, or annotation of food datasets. Ontologies are core components of major data generation projects such as The Encyclopedia of DNA Elements (ENCODE) (5) and the Human Cell Atlas (6). The construction and maintenance of ontologies is a knowledge- and resource-intensive task, carried out by dedicated teams of ontology editors, working alongside the curators who use these ontologies to curate literature and annotate data. Due to the pace of scientific change, the rapid generation of diverse data, the discovery of new concepts, and the diverse needs of a broad range of stakeholders, most ontologies are perpetual works in progress. Many ontologies have thousands, or tens of thousands of terms, and are continuously growing. There is a strong need for tools that help ontology editors fulfill requests for new terms and other changes.

Currently, most ontology editing workflows involve manual entry of multiple pieces of information (also called *axioms*) for each term or class in the ontology. This information includes the unique identifier, a human-readable label, a textual definition, as well as relationships that connect terms to other terms, either in the same ontology or a different ontology (7). For example, the Cell Ontology (CL) (8) term with the ID CL:1001502 has the label "mitral cell", a subClassOf (is-a) relationship to the term "interneuron" (CL:0000099), a "*has soma location*" relationship (9) to the Uberon term "olfactory bulb mitral cell layer" (UBERON:0004186), as well as a textual definition: *The large glutaminergic nerve cells whose dendrites synapse with axons of the olfactory receptor neurons in the glomerular layer of the olfactory bulb, and whose axons pass centrally in the olfactory tract to the olfactory cortex*. Most of this information is entered manually, using either a dedicated ontology development environment such as Protégé (10) or using spreadsheets that are subsequently translated into an ontology using tools like ROBOT (11). In some cases, the assignment of an *is-a* relationship can be automated using OWL reasoning (12), but this relies on the ontology developer specifying logical definitions (a particular kind of axiom) for a subset of terms in advance. This strategy is used widely in multiple different biological ontologies (bio-ontologies), in particular, those involving many compositional terms, resulting in around half of the terms having subclass relationships automatically assigned in this way (13–16).

Except for the use of OWL reasoning to infer *is-a* relationships, the work of creating ontology terms is largely manual. The field of Ontology Learning (OL) aims to use a variety of statistical and Natural Language Processing (NLP) techniques to automatically construct ontologies, but the end results still require significant manual post-processing and manual curation by experts (17), and currently no biological ontologies make use of OL. Newer Machine Learning (ML) techniques such as *link prediction* leverage the graph structure of ontologies to predict new links, but state of the art ontology link prediction algorithms such as rdf2vec (18) and owl2vec* (19) have low accuracy, and these also have yet to be adopted in standard ontology editing workflows.

A new approach that shows promise for helping to automate ontology term curation is instruction-tuned LLMs (20) such as the gpt-4 model that underpins ChatGPT (21). LLMs are highly generalizable tools that can perform a wide range of generative tasks, including extracting structured knowledge from text and generating new text (22,23). One area that has seen widespread adoption of LLMs is software engineering, where it is now common to use tools such as GitHub Copilot (24) that are integrated within software development environments and perform code autocompletion. We have previously noted analogies between software engineering and ontology engineering and have successfully transferred tools and workflows from the former to the latter (25). We are therefore drawn to the question of whether the success of generative AI in software could be applied to ontologies.

Here we describe and evaluate DRAGON-AI, an LLM-backed method for assisting in the task of ontology term *completion*. Given a portion of an ontology term (for example, the label/name, or the definition), the goal is to generate other requisite parts (for example, a textual description, or relationships to other terms). Our method accomplishes this using combinations of latent knowledge encoded in LLMs, knowledge encoded in one or more ontologies, or semi-structured knowledge sources such as GitHub issues. We demonstrate the use of DRAGON-AI to generate both logical relationships and textual definitions over ten different ontologies drawn from the Open Biological and Biomedical Ontologies (OBO) Foundry (26). To evaluate the automated textual definitions, we recruited ontology editors from the OBO community to rank these definitions according to three criteria.

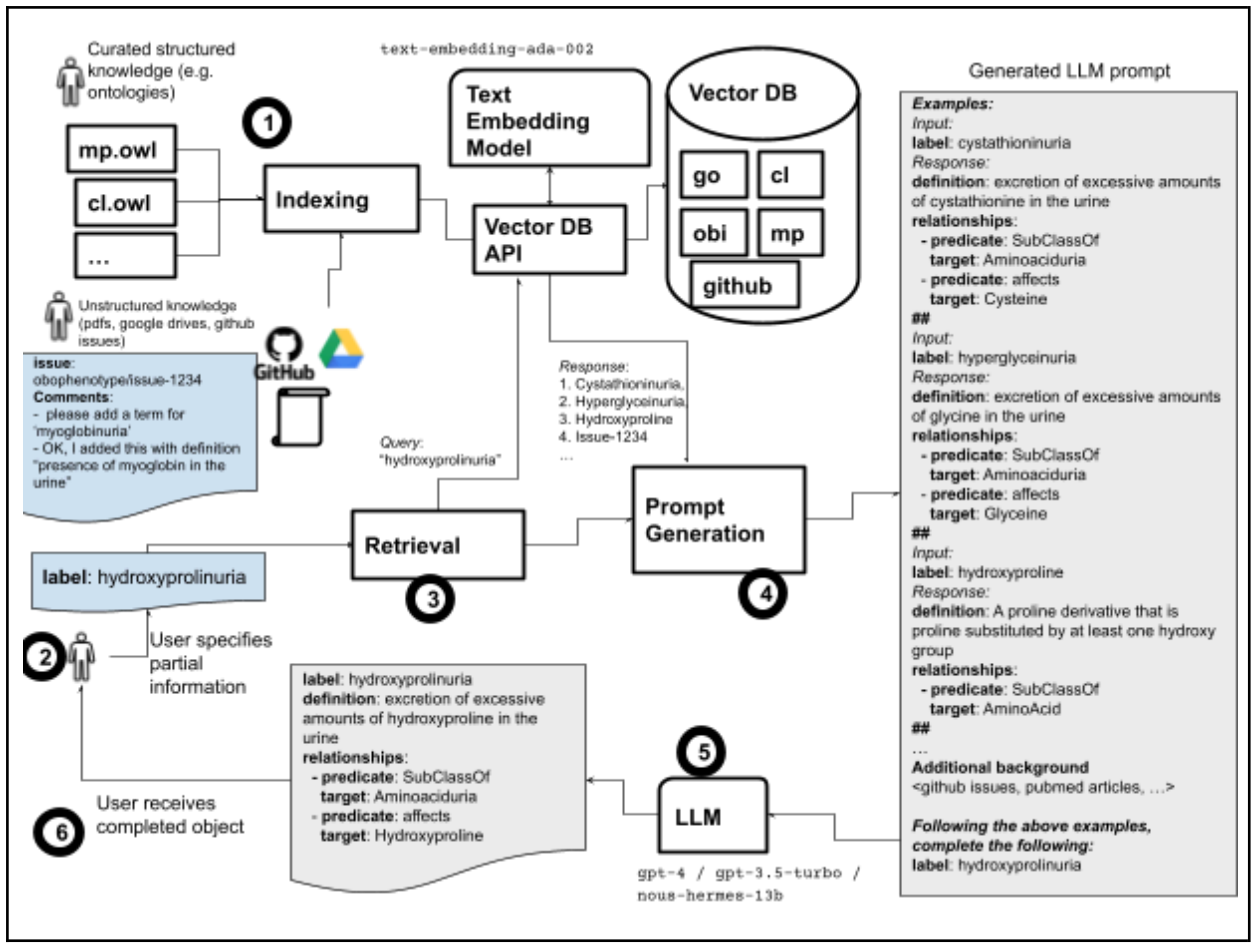

**Figure 1**: **The DRAGON-AI ontology term completion process.** (1) As an initial preprocessing step, knowledge resources (such as ontologies and GitHub issues) are indexed in a vector database. (2) A user provides a partial ontology term object (here, a term with only the label of the desired term "hydroxyprolinuria" is provided). (3) The vector database is queried for similar terms (e.g. cystathioninuria, hydroxyproline) or other relevant pieces of information (e.g. a GitHub issue). (4) A prompt is generated from a template, incorporating the most similar items in the vector database. (5) The prompt is provided as textual input to a LLM, which returns a completed JSON object. Either local or remote LLMs can be used. (6) The parsed object is returned to the user. Note that the above figure uses YAML syntax to represent JSON objects, for the sake of compactness.

We demonstrate that DRAGON-AI is able to achieve meaningful performance on both logical relationship and text generation tasks.

# Implementation

DRAGON-AI is a method that allows for AI-based auto-completion of ontology objects. The input for the method is a partially completed ontology term (for example, just the term label, such as "hydroxyprolinuria"), and the output is an object that has all desired fields populated, including the text definition, logical definition, and relationships.

The procedure is shown in Figure 1. As an initial step, each ontology term and any additional contextual information is translated into a vector embedding, which is used as an index for retrieving relevant terms. Additional contextual information can include the contents of a GitHub issue tracker, which might contain text or semi-structured information of relevance to the request. The main ontology completion step works by first constructing a prompt using relevant contextual information. The prompt is passed as an input to an LLM, and the results are parsed to retrieve the completed term object.

## Indexing ontologies and ontology embeddings

As an initial step, DRAGON-AI will create a vector embedding (27) for each term. Each term is represented as a structured object which is serialized using JSON, following a schema with the following properties:

- `id`: a translated identifier for the term, as described below
- `label`: a string with a human readable label or name for the term
- `definition`: an optional string with a human-readable textual definition
- `relationships`: a list of relationship objects
- `original_id`: the original untranslated identifier
- `logical_definitions`: an optional list of relationship objects

A relationship object has the following properties:

- `predicate`: a translated identifier for the relationship type. For bio-ontologies, this is typically taken from the Relation Ontology (28), or is the subClassOf predicate, for is-a relations
- `target`: a translated identifier for the term that the relationship points to, either in the same ontology, or a different ontology

Ontology terms are typically referred to using non-semantic numeric identifiers (for example, CL:1001502). These can confound LLMs, which have a tendency to hallucinate identifiers. In our initial experiments, we found LLMs tend to perform best if presented with information in the same way that information is presented to humans, presumably as the majority of their training data is in this form. Therefore, we chose to transform all identifiers from a non-semantic numeric form (e.g. CL:1001502) to a symbol represented by the ontology term label in camel case format (e.g. MitralCell). An example is shown in Table 1.

**Table 1. Example JSON structure used in DRAGON-AI**. OBO format syntax is shown at the top, and the corresponding JSON object form shown is below**.** Note that the JSON form omits some information from the OBO Format (e.g. the provenance of the definition).

```
id: CL:1001502
name: mitral cell
is_a: CL:0000099 ! interneuron
relationship: RO:0002100 UBERON:0004186 ! has-soma-location\
olfactory bulb mitral cell layer
definition: "The large glutaminergic nerve cells whose
dendrites synapse with axons of the olfactory receptor
neurons in the glomerular layer of the olfactory bulb, and
whose axons pass centrally in the olfactory tract to the
olfactory cortex" [MP:0009954]
```

```
{
  "id": "MitralCell",
  "original_id": "CL:1001502"
  "relationships": [
   { "predicate": "SubClassOf", "target": "Interneuron" },
   { "predicate": "HasSomaLocation", "target":
"OlfactoryBulbMitralCellLayer" }
  ],
  "definition":  "The large glutaminergic nerve cells whose
dendrites synapse with axons of the olfactory receptor
neurons in the glomerular layer of the olfactory bulb, and
whose axons pass centrally in the olfactory tract to the
olfactory cortex"
}
```

We create a vector embedding for each term by first translating the object to text, and then embedding the text. The text is created by concatenating the label, definition, and relationships. For this study we used the OpenAI *text-embedding-ada-002* text embedding model, accessed via the OpenAI API.

We store objects and their embeddings using the ChromaDB database (29). This allows for efficient queries to retrieve the top *k* matching objects for an input object, using the Hierarchical Navigable Small World graph search algorithm (30).

## Indexing unstructured and semi-structured knowledge

Additional contextual knowledge can be included in DRAGON-AI to inform the term completion process -- for example, publications from PubMed, articles from Wikipedia, or documentation intended for human ontology editors. One of the most important sources of knowledge for ontology terms is the content of GitHub issue trackers, where new term requests and other term change requests are proposed and discussed. Information in these trackers may be free text, or partially structured.

We used the GitHub API to load GitHub issues and store the resulting JSON objects, which are indexed without any specialized pre-processing. The text-serialized form of the GitHub JSON object is used as input for the embeddings. We store these JSON objects separately from the main ontology term objects.

## Prompt generation using Retrieval Augmented Generation

At the core of the DRAGON-AI approach is the generation of a prompt that is passed as input to a LLM. The prompt includes the partial term, and an instruction directing the model to complete the term, filling missing information, and return as a JSON object.

In order to guide the LLM to create a term that is similar in style to existing terms, and to guide the LLM to pick existing terms in relationships, we provide additional context within the prompt. This additional context includes existing relevant terms, provided in the same JSON format as the intended response. Rather than hard-wire this additional context (few-shot learning), we want to provide only the most relevant terms as examples. We use Retrieval Augmented Generation (RAG) (31) as the general approach to retrieve the most relevant information. As a first step, the partial term object provided by the user is used as a query to the ontology terms loaded into the ontology vector index. An embedding is created from the text fields of the object (using the same embedding model as was used to index the ontology), and this is used to query the top *k* results (*k* is 10 by default). These form the *in-context* examples for the prompt. The intent is to retrieve terms that are similar to the intended term to inform the prediction of the completed term; for example, if the query term is "hydroxyprolinuria", then similar terms in the ontology such as "cystathionuria" will be informative.

Each retrieved example forms an input-output training pair which is concatenated directly into the prompt by serializing the JSON object, for example:

> input:
> {"label": "cystathioninuria"}
>
> output:
> {"definition": "excretion of excessive amounts of cystathionine in the urine",
>  "Relationships": [ {"predicate": "subClassOf", "target": "Aminoaciduria"} ] }

To diversify search results, we implement Maximal Marginal Relevance (MMR) (32) in order to re-rank results. This helps with inclusion of terms that inform multiple cross-cutting aspects of the requested term, including terms from other relevant ontologies. For example, if the input is "hydroxyprolinuria" then the highest ranking terms may be other phenotypes involving circulating molecules, but by diversifying search results we also include relevant chemical entities from ChEBI like "hydroxyproline".

Optionally, additional information other than the source ontology can be included in the prompt. This includes GitHub issues, documentation written by and for ontology developers, and PubMed articles. For these sources we also use a RAG method to select only the most semantically similar documents.

Different LLMs have different limits on the combined size of prompt and response. In order to stay within these limits, we reduce the number of in-context examples to the maximum number that still fits within the limit, or the number provided by the user, whichever is greater.

## Prompt passing and result parsing

DRAGON-AI allows for a number of different ways to extract structured information as a response. These include using OpenAI function calls, or using a recursive-descent approach via the SPIRES algorithm (33). For this study we evaluated a pure RAG-based in-context approach, as shown in Figure 1.

This prompt is presented to the LLM, which responds with a serialized JSON object analogous to the in-context examples. This response is parsed using a standard JSON parser, with additional preprocessing to remove extraneous preamble text, and the results are merged with the input object to form the predicted object.

Relationship predictions are further post-processed to remove relationships to non-existent terms in the ontology or imported ontologies. Some of these correspond to meaningful relationships to terms that have yet to be added. In the future, the system may be extended to include a step that fills in missing terms, but the current behavior is to be conservative when predicting relationships.

# Evaluation

We used 10 different ontologies in our evaluation: the Cell Ontology (CL) (8), UBERON, the Gene Ontology (GO), the Human Phenotype Ontology (HP) (34), the Mammalian Phenotype Ontology (MP) (35), The Mondo disease ontology (MONDO), the Environment Ontology (ENVO) (36), the Food Ontology (FOODON), the Ontology of Biomedical Investigations (OBI) (37), and the Ontology of Biological Attributes (OBA) (38). These were selected based on being widely used and impactful and covering a broad range of domains, from basic science through to clinical practice, with representation outside biology (the Environment Ontology and FoodOn). This selection also represents a broad range of ontology development styles, from highly compositional ontologies making extensive use of templated design patterns (OBA) to more individually structured. All selected ontologies make use of Description Logic (DL) axiomatizations, allowing for the use of reasoning to auto-classify the ontology, providing a baseline for comparison. Table 2 shows a summary of which tasks were performed and evaluated on which ontologies.

**Table 2**. **Ontologies and ontology versions used for evaluation.** For each ontology we used the standard release product, except for GO, where we used the go-plus version, which has additional relationships to other ontologies. We took the most recent available version of each ontology, and separated the most recent terms into a test set. The minimum (oldest) date of each term is shown.

| Ontology | Version | Date of oldest term in test set | Terms Tested | Tasks Performed and Evaluated |
| --- | --- | --- | --- | --- |
| CL | 2023-07-20/cl.owl | 2023-01-10 | 50 | all |
| ENVO | 2023-02-13/envo.owl | 2021-05-14 | 50 | all |
| FOODON | 2023-05-03/foodon.owl | 2023-01-01 | 50 | all |
| GO | 2023-07-27/extensions/go-plus.owl | 2023-01-03 | 50 | all |
| HP | 2023-07-21/hp.owl | 2023-01-16 | 50 | relationships, definitions |
| MONDO | 2023-08-02/mondo.owl | 2023-04-01 | 50 | all |
| MP | 2023-08-09/mo.owl | 2023-02-08 | 50 | relationships, definitions |
| OBA | 2023-08-24/oba.owl | 2022-11-26 | 50 | all |
| OBI | 2023-07-25/obi.owl | 2022-12-14 | 50 | relationships |
| UBERON | 2023-07-25/uberon.owl | 2023-01-18 | 40 | all |

We subdivided each ontology into a core ontology plus a testing set of 50 terms. Where possible, we selected test terms from the set of terms that were added to the ontology after

November 2022, to minimize the possibility of test data leakage. This was not possible for ENVO, which has a less frequent release schedule, with the most recent release at the time of analysis being from February 2023, so this ontology included terms added in 2021 and 2022. Uberon also had fewer new terms in 2023, so the test set for this ontology was 40 terms.

We chose three tasks: prediction of (1) relationships, (2) definitions, and (3) logical definitions. For each task, the test set consists of ontology term objects with the field to be predicted masked (other fields such as the ontology term identifier were also masked, as these are another source of training data leakage). For example, to predict relationships, the text objects have only labels and text definitions present.

**Table 3**: **Models evaluated, plus their versions/checkpoints.** The OpenAI training set cutoff dates are based on what is reported on the OpenAI website.

| Model | Checkpoint / Version | Training set cutoff | Access | Description |
| --- | --- | --- | --- | --- |
| gpt-3.5-turbo | 0613 | 2021-09 | API | Proprietary model from OpenAI |
| gpt-4 | 0613 | 2021-09 | API | Proprietary model from OpenAI |
| nous-hermes-13b-ggml | 2023-06 | 2023-02 | Local | Local quantized model fine tuned from llama |

We tested three models (gpt-4, gpt-3.5-turbo, and nous-hermes-13b-ggml) against all ontologies for the three tasks. The first two models are proprietary closed models accessed via an API; the latter model is open, and was executed locally on an M1 MacOS system.

## Relationship prediction evaluation

One of the main challenges in ontology learning is evaluation, since the construction of ontologies involves some subjective decisions, and many different valid representations are possible (39). An additional challenge is that ontologies allow for specification of things at different levels of specificity. For the relationship prediction task, we chose to treat the existing relationships in the ontology as the gold standard, recognizing this may penalize alternative but valid representations.

If a predicted relationship matches a relationship that exists in the ontology, this counts as a true positive. If a predicted relationship is more general than a relationship in the ontology, then we do not count this as a true positive, but instead treat it as a partial false negative (0.5). A relationship (*s, p, o*) is counted as more general if the target node is traversable from the subject node over a combination of is-a (subClassOf) relationship and *p* relationship.

As a baseline, we also include OWL reasoning results using the Elk reasoner (40). This is only applicable to subsumption (SubClassOf) relationships. For each subsumption relationship in the ontology, we remove the relationship and use the reasoner to determine if the relationship is recapitulated. We use the OWLTools (41) tag-entailed-axioms command to do this. As all ontologies use OWL Reasoning as part of their release process, the precision of reasoning, when measured against the released ontology, is 1.0 by definition. However, recall and F1 (42) can be informative to determine breadth of coverage of reasoning.

### Definition prediction evaluation

For the definition prediction task, we could not employ the same strategy as evaluation, as it is very rare for a predicted definition to match the one that was manually authored in the ontology – however, these cannot be counted as false positives as they may still be good definitions. We therefore enlisted ontology editors and curators to conduct a manual assessment of these definitions.

We first aggregated all generated definitions using all models along with the definitions that had previously been manually curated for the test set terms. We assigned each evaluator a task of evaluating a set of definitions by scoring using three different criteria. See supplementary methods for the templates used. The three scoring criteria were:
- *Biological Accuracy*: is the textual definition biologically accurate?
- *Internal consistency*: is the structure and content of the definition consistent with other definitions in the ontology, and with the style guide for that ontology?
- *Overall score*: overall utility of the definition.

For each of these metrics, an ordinal scale of 1–5 was used, with 1 being the worst, 3 being acceptable, and 5 being the best. Assigning a consistency score was optional. Evaluators could also choose to use the same score for accuracy and overall score. Additionally, the evaluator could opt to provide a confidence score for their ranking, also on a score ranging from 1 (low confidence) to 5 (high confidence). We provided a notes column to allow for additional qualitative analysis of the results.

At least two evaluators were assigned to each ontology. Evaluators received individualized spreadsheets and were blinded from the source of the definition. They worked independently, and did not see the results of other evaluators until their task was completed. Evaluators were also asked to provide a retrospective qualitative evaluation of the process, which we include in the discussion section.

## Execution

Our workflow is reproducible through our GitHub repository (43), also archived on Zenodo (44). A Makefile is used to orchestrate extraction of ontologies, splitting test sets, loading into a vector database, and performing predictions. A collection of Jupyter Notebooks is used to evaluate and analyze the results.

# Results

## AI-generated relationships have high precision but moderate recall

For each generated term across all 10 ontologies, we evaluated the generated relationships by comparing them to existing relationships in the ontology. We subdivided this evaluation into two parts: (1) evaluating only subsuming parents (is-a/SubClassOf relationships) and (2) evaluating all relationships (making use of heterogeneous relationship types). We compared AI-generated relationships against the use of DL reasoning using the Elk reasoner.

The aggregated results of the evaluation are summarized in Table 4. In all cases, the best performing model for use with DRAGON-AI is gpt-4. On the SubClassOf subtask, the best performing model has high precision (0.889), which is comparable with, but less precise than, using DL reasoning (which by its nature always has maximal precision). On this subtask, DRAGON-AI has better recall and F1 than using reasoning. For the heterogeneous relationship subtask, the overall scores are lower (0.797 for precision), but still indicate strong performance. Note that this subtask is outside the capabilities of DL reasoning.

**Table 4: DRAGON-AI results for relationship prediction task.** We partition into two subtasks: filtered for SubClassOf, and filtered for all relationship types (heterogeneous relationship predictions). OWL Reasoning results included as baseline for SubClassOf. Note that by definition OWL reasoning is always completely precise as all entailments follow from existing axioms.

|         |                | SubClassOf Task |        |       | All Relationship Types Task |        |       |
|---------|----------------|----------------:|-------:|------:|----------------------------:|-------:|------:|
| method  | model          | precision       | recall | F1    | precision                   | recall | F1    |
| DRAGON  | gpt-3.5-turbo  | 0.831           | 0.352  | 0.494 | 0.746                       | 0.392  | 0.514 |
| DRAGON  | gpt-4          | **0.889**       | **0.44** | **0.588** | **0.797**               | **0.456** | **0.58** |
| DRAGON  | nous-hermes-13b| 0.68            | 0.273  | 0.39  | 0.597                       | 0.292  | 0.392 |
| Reasoner| n/a            | *1.0**          | 0.337  | 0.504 | n/a                         | n/a    | n/a   |

We also observed that different ontologies may be more or less amenable to relationship prediction, as shown in Figure 2. However, note that the test set distribution may not be reflective of the overall distribution in the ontology, as we limited testing to new terms only.

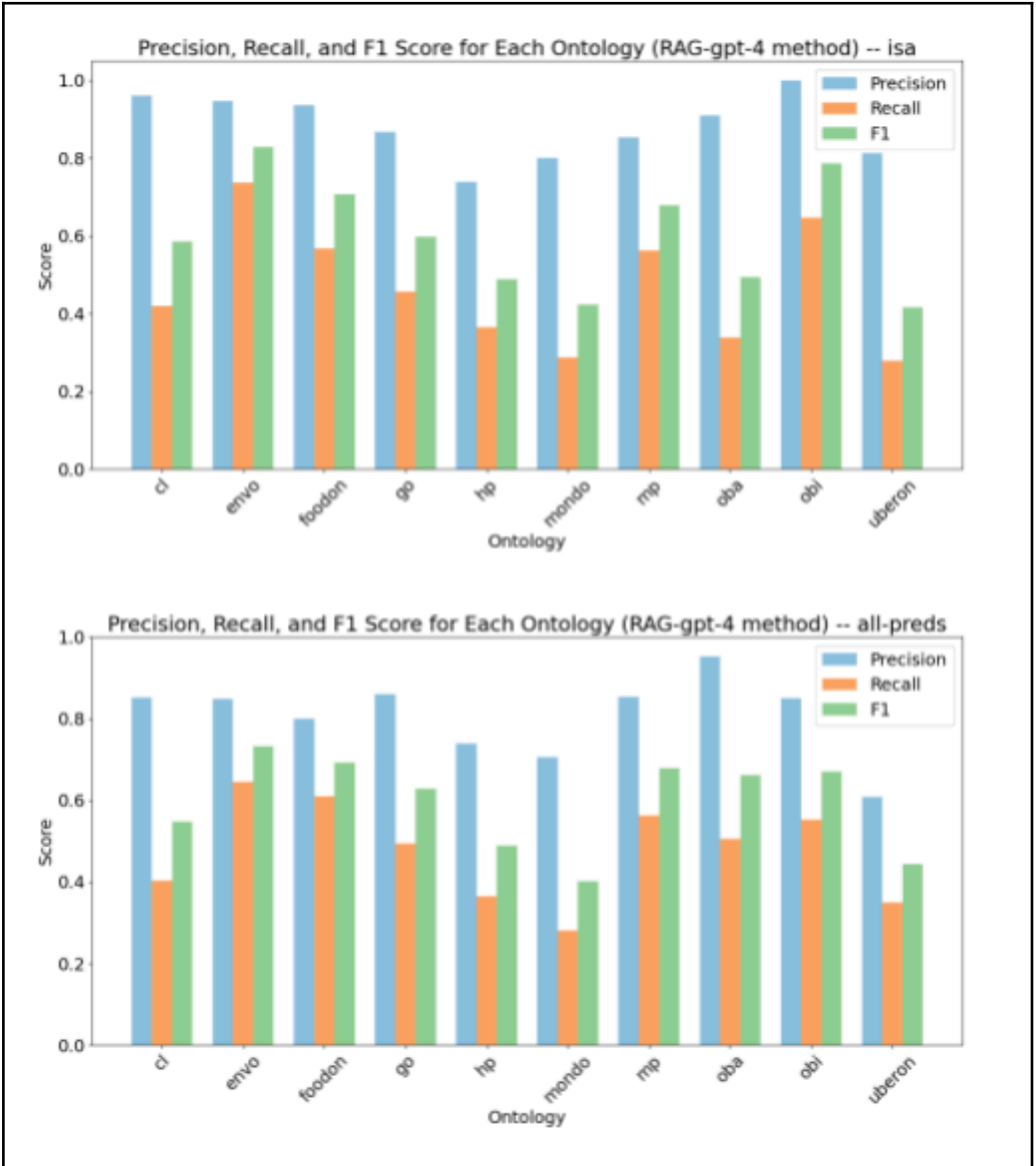

**Figure 2**: Metrics for relation prediction across 10 ontologies (gpt-4 results only, filtered for SubClassOf/is-a and all relationship types).

## AI-generated definitions score well, but less than existing definitions

For all ontologies, we generated text definitions for each term in the test set, providing only the label and relationships, plus logical definitions if present. These generated definitions were evaluated by curators and ontology editors and scored according to different criteria. Intraclass Correlation Coefficients (ICC) revealed curators to be largely consistent in their scoring of definitions (ICC scores of 0.799, 0.737, 0.770 for accuracy, consistency, and overall score).

**Table 5**. **DRAGON-AI performance on definition generation task**. A comparison of base

performance of DRAGON on definition generation when compared with existing editor-provided definitions Evaluator scores shown for three score categories (accuracy, consistency, and overall score). Evaluators evaluated definitions generated by three different models, alongside existing ontology definitions. Evaluators were not shown the source of definitions until after evaluation.

| method | model name | accuracy | score | consistency |
|---|---|---|---|---|
| DRAGON | gpt-3.5-turbo | *4.058* | *3.632* | *3.735* |
| DRAGON | gpt-4 | 3.97 | 3.567 | 3.689 |
| DRAGON | nous-hermes-13b | 3.776 | 3.389 | 3.566 |
| curator | human | **4.326** | **4.069** | **4.13** |

Overall, definitions authored by human curators scored highest on all three metrics (Table 5). DRAGON performed acceptably (consistently above a grade of 3, which was considered acceptable) regardless of the underlying model, with gpt models outperforming the only open model evaluated (nous-hermes-13b). The performance gap between curated definitions and generated definitions is statistically significant for all score types. The gap between gpt-3.5-turbo or gpt-4 and the open model was also statistically significant. However, the gap between gpt-3.5-turbo and gpt-4 was not significant.

The results of the manual evaluation are also available on HuggingFace (45).

## Experts are more likely to detect flaws in AI-generated definitions

The difference between the manually authored definitions and the best AI generated definitions is statistically significant, yet moderate in effect. We hypothesized that this difference would decrease as the evaluator confidence decreases – i.e. less confident evaluators would be less able to discriminate between a good definition and plausible yet flawed definition.

When we plot the performance gap between the best performing model and human curation, we can see a clear correlation between performance gap and confidence, with the lowest confidence showing no discrimination between model-generated and human curation. The correlation is highly significant (Pearson correlation coefficient 0.973).

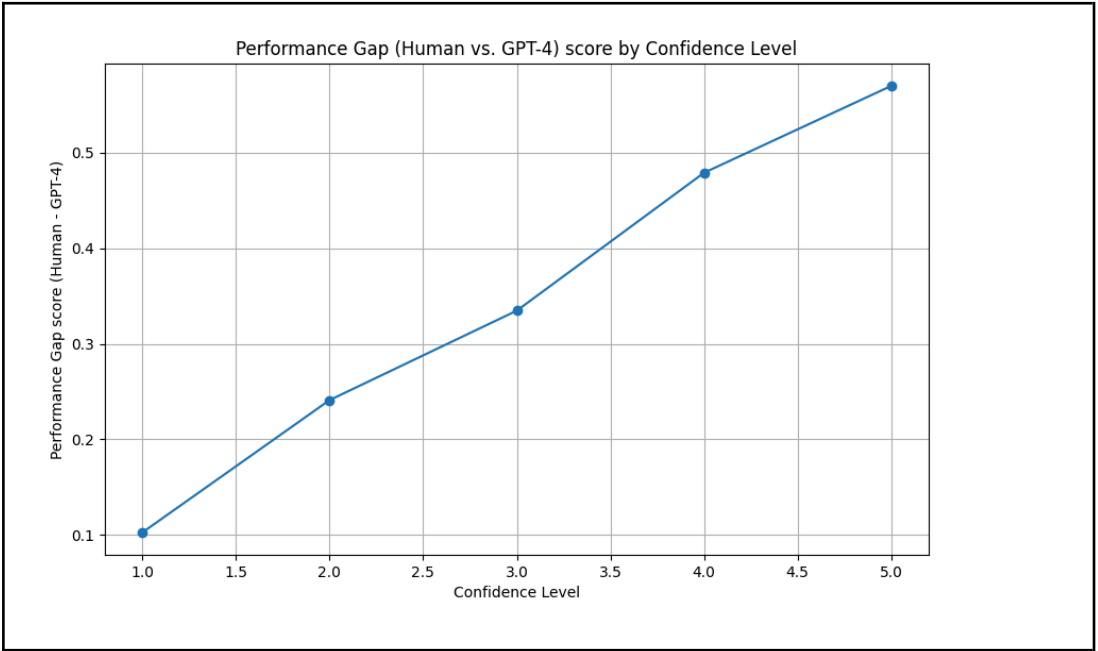

**Figure 3**: **Performance gap vs confidence level.** If an evaluator lacked confidence in their assessment (lower confidence level), they were more likely to assign an LLM-generated definition a comparable score to a human curated one. As the evaluator confidence increases, the evaluator is more likely to rank the LLM-generated definition lower than the human one.

## DRAGON-AI can read and interpret GitHub issues to improve performance

We investigated whether providing background knowledge from GitHub issue trackers would improve the quality of generated definitions. This information is generally provided in text or semi-structured form. We applied the method *RAG+github*, in which RAG is used to retrieve both the most relevant ontology terms and the most relevant GitHub issues, and both are included in the prompt. We restricted this analysis to two models (gpt-4 and gpt-3.5-turbo) and three ontologies (CL, UBERON, and ENVO).

**Table 6**. **Comparison of scores when GitHub issues are included as background knowledge.**

| method | model name | accuracy | score | consistency |
| --- | --- | --- | --- | --- |
| DRAGON | gpt-3.5-turbo | 4.067 | 3.626 | 3.709 |
| DRAGON+gh | gpt-3.5-turbo | 4.182 | 3.717 | 3.733 |
| DRAGON | gpt-4 | 4.041 | 3.608 | 3.754 |
| DRAGON+gh | gpt-4 | 4.241 | 3.805 | 3.893 |
| curator | human | **4.439** | **4.158** | 4.182 |

Including the GitHub issues improved performance of all models, although performance was still beneath manually authored definitions (Table 6). The difference between RAG with and without GitHub is statistically significant for both accuracy and score.

Overall this indicates that generative AI can make use of sources of information intended primarily for humans as a part of their term creation workflow.

## Logical definitions can be generated with high accuracy in some ontologies

We evaluated the ability of DRAGON-AI to generate logical definitions across four different ontologies. Only a subset of ontologies were used, as other ontologies did not have a sufficient number of logical definitions in newly added terms to test against, or logical definitions did not conform to the simple genus-differentia form.

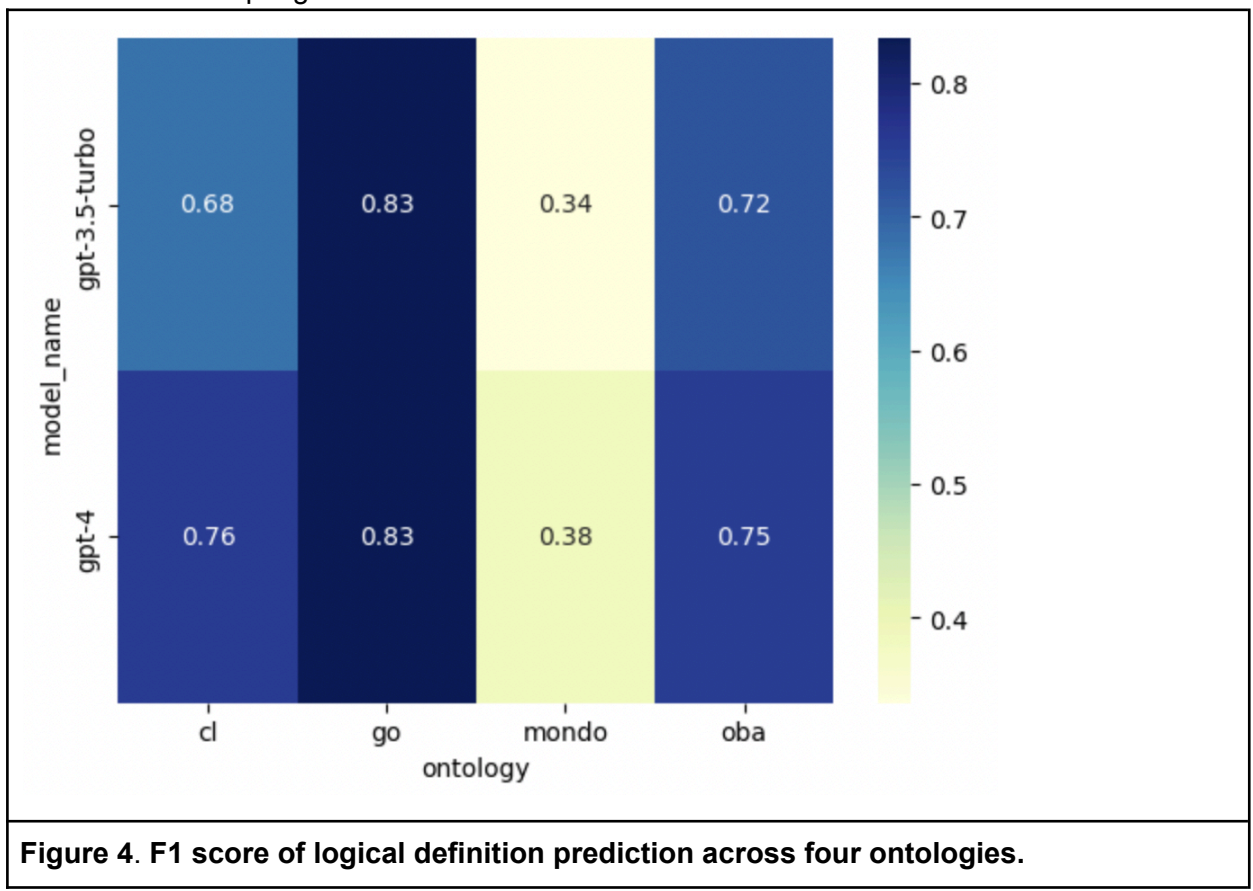

**Figure 4**. **F1 score of logical definition prediction across four ontologies.**

The results are shown in Figure 4, demonstrating wide variability.

# Discussion

## Generative AI shows promising capabilities to assist in ontology editing workflows, but should be used with caution

Our results demonstrate the feasibility of incorporating generative AI into ontology development workflows. For relationship generation, when we compare with existing ontology relationships, we demonstrate high precision, and moderate recall/F1. This indicates that results are generally correct, but may be incomplete. Even when AI results do not conform to asserted relationships in the ontology, they frequently represent a valid perspective that could be incorporated. For definition generation, AI-authored definitions rank close to yet lower than human-authored ones. The DRAGON-AI system is also able to leverage textual information from other sources to enhance its results. Additionally, DRAGON-AI is able to leverage additional textual sources of information such as GitHub issues.

Note that we do not expect AI-generated axioms to be perfect in order for them to be useful. We envision DRAGON-AI being used as part of an autocomplete system within existing ontology development environments like Protégé, or in tabular editing environments used in conjunction with tools such as ROBOT. Here the editor can be presented with suggested axioms to add, based on partial information they have entered, with the ability to easily accept, reject, or modify AI-generated suggestions, and ultimately even the ability to interact with the system using natural language in order to hone results. This kind of autocomplete paradigm is already widely used in software development environments through tools such as GitHub Copilot. Copilot has been widely adopted by software developers (over one million paid subscribers), with most users self-reporting increased productivity (46). This is despite the fact that Copilot suggestions are only accepted 30% of the time, in contrast with the precision of 82% we have achieved for ontology completion.

However, any such tool should be used with caution. For software development, some studies have shown that while AI tools can boost productivity, they can also be a liability for novice users (47). Our results also showed that novice editors are more likely to be "tricked" by the AI. We demonstrated that if an evaluator had lower confidence in a domain, they were more likely to accept a generated definition on face value, even if incorrect. Evaluators with more experience in a domain are more likely to spot subtle problems with generated terms. We informally call this the "gaslighting effect", and in fact a number of evaluators commented on the fact that they thought they were being "gaslit" by the AI. This is similar to a previously observed phenomenon of LLMs "sandbagging" their users (48). This means that AI should be used with caution, particularly in the hands of less experienced ontology editors. Of course, it is also important to point out that ontology developers can make mistakes without AI, and all ontologies should employ appropriate centralized QA/QC measures. In fact, assisting with whole-ontology QA/QC represents a potential useful future application of AI.

Overall, our goal is to enhance the experience of ontology editors and improve productivity. A recent study indicates that generative AI can help restructure tasks towards idea generation and away from tedious repetitive tasks (49). Our vision for DRAGON-AI is a tool that allows ontology editors and curators to employ their deep understanding of a domain to efficiently translate that knowledge, minimizing tedious tasks such as copying information from reference sources.

## Challenges in evaluating ontologies by LLMs

There are a number of challenges in evaluating the effectiveness of LLMs in ontology generation, including test data leakage and the inherent subjectivity of ontology construction. Test data leakage occurs when the LLM training sets are contaminated with benchmark data. This is common, due to the fact they are trained on internet-sized corpora (50,51). Most LLMs have effectively memorized most public bio-ontologies. To minimize the possibility of test data leakage, we used only terms that had been added to ontologies after the training data cutoff of the major GPT models. However, this limited the size of the test corpus, and may have potentially limited the generalizability of the results, at least within ontologies (for many ontologies, the terms entered in a one year period may not be representative of the overall content of the ontology). It also makes it hard to evaluate the effectiveness of LLMs on *de novo* ontology construction, since the RAG approach makes heavy use of existing terms. Ontologies such as the GO and the Cell Ontology have existed for over two decades, and are the combined efforts of a massive number of editors, curators, and domain experts. We consider it a strength of the DRAGON-AI method that it is able to leverage this prior work via RAG when generating new terms; we consider this task relevant to the day to day efforts of biological ontology developers. However, it also means we did not address the question of how well the approach would perform on constructing an ontology from scratch, or from an early state. We are currently exploring the use of DRAGON-AI in the creation of *de novo* ontologies in the environmental domain. We have created an experimental ontology of over a thousand environmental variables and parameters for use in earth system simulation (52).

It is also important to emphasize that the success of AI methods on the ontologies we evaluated depends largely on the previous work of hundreds of ontology editors and curators working over decades. Ontologies are included within the datasets used in LLM pre-training, so the LLM already comes pre-equipped to recognize common patterns employed within these ontologies.

The paradigm of using terms added after the training date cutoff for evaluation is likely to become even less effective over time, as it becomes easier to update models with new data. The version of gpt-4 and gpt-3.5-turbo we used in this evaluation had a training date cutoff of September 2021, but the latest versions of these models have far more recent cutoffs. The open models we used also had recent cutoffs. Training a model from scratch with an older cutoff for evaluation purposes is simply not feasible due to the massive costs involved in pre-training. It is therefore vital that we invest in efforts to evaluate ontology generation on new domains using terms that have not been used in pre-training.

The other challenge involves the inherent subjectivity of ontology construction; there are different ways to represent the same thing. For our relationship prediction task we took the terms that were created by ontology editors as the gold standard, but the predicted relationships that do not match these terms are not necessarily incorrect. When we manually examined the relationships counted as false positives, many were alternate representations, or simply relationships missing from the ontology (see supplementary material).

To overcome both these challenges, investing more in expert human evaluation is essential.

# Future directions

## Customizing RAG

One of the strengths of DRAGON-AI is in-context learning from existing terms in the ontology. However, not all terms in an ontology are of equal quality. Ontologies that have been developed over long time frames may include "legacy" terms that do not serve as good exemplars. We therefore plan to extend the approach to allow the user to influence the ranking of terms returned by RAG, for example by prioritizing newer terms (presumably these are more reflective of current best practice in the ontology), or allowing the use of metadata that marks certain terms as being good examples of best practice for particular kinds of terms.

## Incorporation into ontology editing environments

In order for AI methods to be successful, they need to be seamlessly integrated into ontology editing workflows. For future development, we are considering a number of different potential workflows.

The first workflow would be integrating AI methods into existing ontology development tools such as Protégé via a plugin. The plugin would function analogous to AI-based code completion in software development environments; the editor would create a new term, provide a label, and the plugin would suggest a completed term, which the editor could either accept outright, accept and then modify, or reject.

The second workflow would be integration into a tabular editing environment, where the editable tables are used as part of a tabular template-driven workflow, supported by a tool such as ROBOT templates (11), Dead Simple Design Patterns (DOSDPs) (53), OTTR templates (54), or LinkML (55).

The third workflow would be to design a new kind of user interface that reflects a potential new role for ontology editors, with less emphasis on data entry and more on high level specification of requirements and evaluation and honing of AI generated content. Here the interface may focus on text-oriented interactions, as in ChatGPT, coupled with easy ways to guide the AI. To this end, we have commenced work on a general-purpose AI-driven curation Integrated development environment (IDE) called CurateGPT (56). All of the workflows evaluated in this manuscript are supported in the current UI. However, a number of challenges need to be overcome to make this IDE usable. Some of these challenges involve the current low latency of LLM prompt completion; others involve making the interface more user friendly, which will require extensive feedback and iterative testing from ontology editors and curators.

A fourth workflow is integration directly into LLM chat interfaces. One such mechanism is the "GPTs" feature of ChatGPT. We have recently developed a ROBOT template GPT helper (57) and used this in the development of the Artificial Intelligence Ontology (AIO) (58).

Regardless of which interface paradigm is followed, we believe the most important functionality is a simple and intuitive way to accept, reject, or modify AI generated suggestions, as well as recording these responses, in order to continually improve the system.

## Automated methods to validate generative AI results

In order to increase performance, and in particular, quality and reliability of results, we are exploring a two-pronged approach to including automated validation as part of the DRAGON-AI workflow.

The first approach is to couple DRAGON-AI with an OWL reasoner: this will allow for filtering of redundant relationships, as well as inference of implicit relationships. Note that the recall of OWL reasoning increases with the degree of axiomatization in ontologies, and many ontologies are under-axiomatized. Here we propose an approach involving generation of additional constraint style axioms, such as disjointness axioms, that will allow for OWL reasoners to detect errors in generated terms. It should also be possible for DRAGON-AI to both generate and populate ontology design patterns.

The second approach involves using methods such as RAG to try to find evidence for generated statements in the literature. In many ontologies, statements are accompanied by provenance information, such as bibliographic references or references to sources such as Wikipedia. In future studies we aim to combine DRAGON-AI with the Evidence Agent in CurateGPT (56), which is able to retrieve relevant references and apply as evidence. This could also be used as an additional filter.

## Support for additional workflows

In this paper we demonstrate the use of DRAGON-AI for term generation. However, for many ontologies, new term requests only constitute one part of the overall workflow. Maintenance and correction of existing terms can also be resource-intensive, especially for ontologies with tens of thousands of terms collected over decades. Often it is necessary to "refactor" ontologies, where large numbers of terms are modified together, for example, as part of an overhaul of how a particular area of biology is reflected. We aim to extend DRAGON-AI to support these additional workflows, and, in particular, to make use of rich information already present in many GitHub issue trackers that couple requested changes with enacted changes, in order to build something more like an autonomous agent that is able to work through large numbers of requested changes specified in free text, interacting with domain experts and ontology editors through conversational mechanisms.

# Conclusions

Building and maintaining ontologies is time-consuming and requires substantial human expertise. DRAGON-AI demonstrates the potential of generative AI approaches, in conjunction with human oversight, to facilitate these tasks. DRAGON-AI can draw on structured knowledge from multiple ontologies, as well as textual sources including GitHub issues that request ontology changes.

We tested DRAGON-AI on three ontology editing tasks: prediction of relationships, term definitions, and logical definitions. Its performance was evaluated by 24 ontology editors and curators who worked independently of each other; each ontology was reviewed by at least 2 evaluators. Based on these evaluations, we found that AI-generated relationships had high precision but moderate recall, suggesting that they were generally correct but incomplete. The AI-generated term definitions were found to be decent but not as good as human-generated definitions. One interesting finding was that the more experienced the evaluator was, the more difference they tended to perceive in the quality of the human-generated vs. AI-generated definitions.

We are investigating ways to incorporate generative AI approaches into existing ontology development workflows. Our ultimate goal is not to replace human ontology editors, but rather to augment their deep domain expertise with tools that minimize tedious, repetitive tasks and make ontology creation and editing more efficient without sacrificing accuracy.

# Availability and requirements

Project name: DRAGON-AI
Project home page: https://github.com/monarch-initiative/curate-gpt (DRAGON-AI is implemented as part of the Curate-GPT suite of tools).
Operating system(s): Platform independent
Programming language: Python
Other requirements: N/A
License: BSD-3
Any restrictions to use by non-academics: none; free to reuse and modify

# List of abbreviations

AI: Artificial Intelligence
API: Application Programming Interface
CL: Cell Ontology
DL: Description Logic
DRAGON-AI: Dynamic Retrieval Augmented Generation of Ontologies using Artificial Intelligence
ENVO: Environment Ontology

GO: Gene Ontology
HP: Human Phenotype Ontology
ICC: Intraclass Correlation Coefficients
IDE: Integrated development environment
JSON: JavaScript Object Notation
LLM: Large Language Model
ML: Machine Learning
MMR: Maximal Marginal Relevance
MP: Mammalian Phenotype Ontology
NLP: Natural Language Processing
OBA: Ontology of Biological Attributes
OBI: Ontology of Biomedical Investigations
OBO: Open Biological and Biomedical Ontologies
OL: Ontology Learning
RAG: Retrieval Augmented Generation
SPIRES: Structured Prompt Interrogation and Recursive Extraction of Semantics

# Declarations

### Ethics approval and consent to participate

Not applicable

### Consent for publication

Not applicable

### Availability of data and materials

The datasets generated and/or analyzed in the current study are available in https://github.com/monarch-initiative/; a stable version is archived in Zenodo (44). The workflow and Jupyter notebooks for the evaluation in this paper can be found at https://github.com/monarch-initiative/gpt-ontology-completion-analysis.

### Competing interests

MAH is a founder of Alamya Health. The other authors declare that they have no competing interests.

### Funding

This work was supported by the National Institutes of Health National Human Genome Research Institute [HG010860, HG012212, HG010859]; National Institutes of Health Office of the Director [R24 OD011883]; and the Director, Office of Science, Office of Basic Energy Sciences, of the US Department of Energy [DE-AC0205CH11231 to CJM, NLH, MJ, and JPR].

TR was funded by the National Science Foundation (NSF) under grant number OAC-2112606 (ICICLE program). TL was supported by a grant from the São Paulo Research Foundation (#19/26284-1). VW was supported by a grant from Wellcome Trust [218236/Z/19/Z]. We also gratefully acknowledge Bosch Research for their support of this research project.

## Authors' contributions

ST and CJM led the research and the writing of this manuscript. MPJ, SON, JTR, and CJM contributed to the software used in this study. NLH, MCMT, and MAH provided project support. NLH significantly edited the manuscript. ST and all other authors participated in the method evaluation as expert biocurators. All authors read and approved the final manuscript.

## Acknowledgements

Support for title page creation and format was provided by AuthorArranger, a tool developed at the National Cancer Institute.

# Supplementary Material
## S1. Additional evaluation of predicted relationships

On examination of the false positives from the relationship generation task, we saw many cases where the AI generated relationships did not match what had been placed in the ontology by the ontology editor, but were nevertheless arguably correct.

In some cases, predictions were valid but less precise than the asserted term.

An example of this was the term "subiliac lymph node" in Uberon, which represents a non-human lymph node. The predicted relationships were subClassOf 'lymph node', but the expected answer in the ontology was 'parietal pelvic lymph node'. The prediction is an anatomically valid relationship, but not as precise as the one in the ontology. We treat these as false negatives, but score this as 0.5 of a false negative

In other cases, the predictions yielded additional more specific yet redundant information. For example, 'long bone cartilage element' is asserted in Uberon to be a subtype of "cartilage element" and a part of "long bone". DRAGON-AI predicts both of these, as well as an additional relationship "composed primarily of 'cartilage tissue'". This is correct, but is redundant since this is inferred from the parent term. But because we score strictly based on recapitulating the precise relationships, this additional prediction is treated as a false negative, reducing the F1 for this term to 0.66, despite it being formally precise and accurate.

A converse case is "epithelium of lobular bronchiole" which was predicted to be a subtype of "epithelium" and a part of the "lobular bronchiole". In the release version of the ontology the supertype is "epithelium of bronchiole". However, the DRAGON-AI suggestion is consistent with what the ontology editor would assert, leaving the reasoner to infer the more specific supertype.

In some cases DRAGON-AI will "hallucinate" parent terms that do not exist that are good candidates for inclusion in the ontology. For example, in the Uberon task, the term "thoracolumbar junction", the existing is-a parent is "organ part", whereas the suggested one was a non-existent term "body junction". Arguably this suggested parent is more informative and a correct grouping term.

In other cases, DRAGON-AI will correctly predict missing relationships in the ontology – for example, the term "right supraclavicular lymph node" did not have any laterality relationships asserted, as is normally standard practice in Uberon. DRAGON-AI correctly predicted the is-a parent ("supraclavicular lymph node") and additionally predicted an "in right side of appendicular skeleton" relationship; this suggestion is consistent with the textual definition, existing ontology patterns, and is anatomically correct. However, the accuracy for this term was scored at 0.5, since the ontology was missing this relationship.

## S2. Subsumption prediction beats state of the art knowledge graph embedding

We compared results with owl2vec*(19), the state of the art Knowledge-Graph Embedding (KGE) based technique. Owl2vec only predicts is-a (subsumption between named classes), so we first filtered our predicted relationships to include only is-a relationships.

Owl2vec returns a ranked list of predictions, as opposed to DRAGON-AI which provided a crisp list (i.e., boolean yes/no with no rankings). For comparison purposes, we treat the top ranking owl2vec prediction (i.e. what is reported as Hits@1 in the paper) as the main prediction.

The owl2vec* method has been tested over FoodOn and GO, so we compared our results with these. The reported Hits@1 for FoodOn and GO are 0.143 and 0.076 respectively. Owl2vec* beats other KGE methods: e.g. rdf2vec scores 0.053 and 0.017 on these ontologies respectively.

For DRAGON, the respective precision scores for DRAGON with gpt-4 are 0.941 and 0.886, which improves considerably over state-of-the-art KGE methods (represented in Table 3 by the best-performing KGE method, owl2vec). Even using the lowest performing model, nous-hermes-13b, DRAGON still outperforms all KGE methods (see Supplementary Table 1).

**Supplementary Table 1**. **Comparison of DRAGON with state of the art KGE methods on is-a (subsumption) relationship prediction task.**

| method | foodon | go |
|---|---|---|
| rdf2vec | 0.053 | 0.017 |
| owl2vec* | 0.143 | 0.076 |
| RAG-nous-hermes-13b | 0.81 | 0.583 |
| RAG-gpt-3.5-turbo | 0.821 | **0.95** |
| RAG-gpt-4 | **0.935** | 0.867 |

## S3: Effects of chain of thought reasoning

We investigated a number of different prompting strategies, including a variant of chain-of-thought, where we first asked the model to generate a de-novo description of the term (effectively using its own latent "ontology"), and then feeding that description back in as part of the context in the RAG prompt. Overall this had mixed effects on definition generation (Supplementary Table 2), and paradoxically lowered gpt-4 performance on relationship generation and boosted gpt-3.5-turbo generation on the same task.

**Supplementary Table 2**. **Effects of including auto-generated background on definition generation**. The effects of including auto-generated background on definition generation are moderate and inconsistent.

| method | model_name | accuracy | score | consistency |
|---:|---:|---:|---:|---:|
| RAG+background | gpt-3.5-turbo | **4.105** | 3.435 | 3.349 |
| RAG | gpt-3.5-turbo | 4.018 | **3.598** | **3.7** |
| RAG+background | gpt-4 | **4.1** | 3.237 | 3.072 |
| RAG | gpt-4 | 3.945 | **3.543** | **3.664** |
| curator | human | **4.332** | **4.06** | **4.107** |

**Supplementary Table 3**. **Effects of including auto-generated background information on relationship generation.** Use of background moderately boosts gpt-3.5-turbo performance and conversely has a negative effect on gpt-4 performance.

| model | method | precision | recall | F1 |
|---:|---:|---:|---:|---:|
| gpt-3.5-turbo | RAG+background | **0.782** | **0.409** | **0.537** |
| | RAG | 0.746 | 0.392 | 0.514 |
| gpt-4 | RAG+background | 0.726 | 0.432 | 0.541 |
| | RAG | **0.797** | **0.456** | **0.58** |

## S4: Automated summary of evaluator notes

We collected all notes from all evaluations, and created a CSV containing the term name, definition, and notes, and used an online AI tool (claude.ai) to summarize general patterns. The results are supplied verbatim below:

Based on reviewing the evaluator notes, some general patterns in the definition evaluations include:

1. Issues with following the Aristotelian definitional structure (genus + differentiae) preferred for definitions in ontologies like CL and FoodOn. Many definitions were missing the genus term, started with the term being defined, lacked sufficient differentiating detail, or contained lengthy extra descriptions better suited for comments.

2. Definitions being overly complex, wordy, or containing unnecessarily detailed technical and biological information. More concise language was often recommended.

3. Circular definitions where the term being defined appeared in its own definition. Explicit definitions of key concepts were needed to avoid circularity.

4. Missing specifics related to the term label itself in the definition, such as failing to mention a key quality like "raw" for a food product.

5. Problems with consistency of style and word choice compared to existing term definitions, such as capitalization, tense, plural vs singular, etc. Standardizing the language used was advised.

6. Issues with accuracy of the biological information provided, including incorrect or narrow characterizations of anatomical entities, molecular processes, etc. Evaluators noted when further research would be needed to confirm definition veracity.

In summary, the main patterns were definitions needing improvement in structure/formatting, conciseness, avoidance of circularity, faithfulness to the term label, consistency with existing definitions, and accuracy of the biological details provided. Following ontology guidelines and best practices for textual definitions was encouraged.